\documentclass[runningheads]{llncs}

\usepackage{eccv}

\usepackage{eccvabbrv}

\usepackage{amsmath}
\usepackage{amssymb}
\usepackage{graphicx}
\usepackage{wrapfig}
\usepackage{booktabs}
\usepackage{xcolor}
\usepackage{placeins} 
\usepackage[accsupp]{axessibility}  

\usepackage{hyperref}

\usepackage{orcidlink}

\begin{document}

\title{ComplexMimic: Human–Scene Interaction Imitation in Complex 3D Environments} 

\titlerunning{ComplexMimic}


\author{ Lu Pan\inst{1,2}\orcidlink{0009-0001-9909-3563}
\and 
Hongwei Zhao\inst{1,2}\orcidlink{0000-0003-2795-8932}}

\authorrunning{L. Pan and H. Zhao}

\institute{College of Computer Science and Technology, Jilin University, China \and Key Laboratory of Symbolic Computation and Knowledge Engineering of Ministry of Education, Jilin University, China
\\ \email{panlutag@gmail.com, zhaohw@jlu.edu.cn}
}

\maketitle

\begin{abstract}
Physics-based Human-Scene Interaction (HSI) imitation learning is crucial for embodied intelligence as it bridges the gap between kinematic 3D motions and real-world dynamics. However, most existing methods focus on simplified scene settings, leaving complex environments largely unexplored, which limits their applicability in real-world scenarios. In this paper, we focus on HSI mimicry in complex environments. Under this complex setting, we observe an inherent trade-off between successfully performing interaction and maintaining natural, physically plausible motions. To address this challenge, we propose ComplexMimic, a framework that reconstructs diverse HSI by interpreting imperfect MoCap data. First, we introduce a Dual Flow Strategy, which learns two complementary experts: an imitation expert for accurate motion tracking and an interaction expert for collision-aware adaptation in complex scenes. Second, naive multi-expert distillation, which treats all experts equally, often under-samples challenging behaviors, limiting effective learning. To mitigate this issue, we propose a difficulty-aware distillation strategy that adaptively weights supervision and prioritizes hard-yet-learnable trajectories guided by failure statistics and learning progress signals. Extensive experiments on three benchmark datasets demonstrate that our approach outperforms current state-of-the-art methods. Our implementation is available at
\url{https://github.com/LuPan23/ComplexMimic}.

\end{abstract}

\section{Introduction}
\label{intro}
\begin{figure}[t] 
    \centering 
    \includegraphics[width=1\textwidth, trim={0 0 0.1cm 0}, clip]{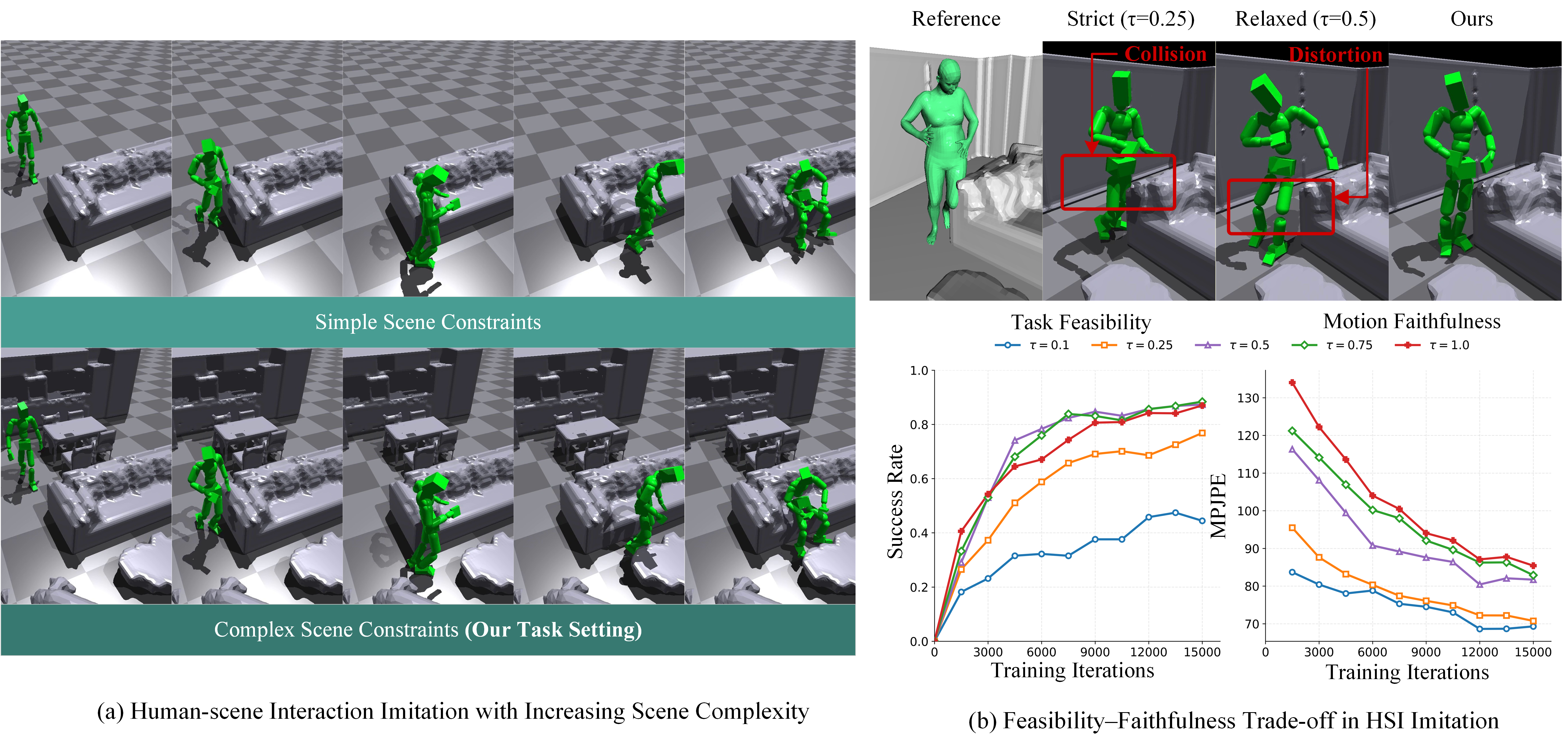}
    \caption{(a) Prior humanoid imitation-learning studies \cite{liu2024mimicking} typically focus on scene-free or overly simplified environments (upper row), whereas our work targets realistic HSI imitation in complex 3D scenes (lower row). (b) Feasibility–faithfulness trade-off in HSI imitation. Varying the early termination threshold $\tau$ during training reveals a clear trade-off: larger $\tau$ improves task feasibility (higher success rate) but reduces motion fidelity (higher MPJPE), while smaller $\tau$ enforces faithful motion at the cost of task success.}

\label{fig:figure_0_motivation_illustration} 
\end{figure}
Humanoid whole-body control via imitation learning plays a crucial role in embodied intelligence and physically plausible character animation~\cite{yu2025skillmimic,peng2018deepmimic,xu2025intermimic,wang2025experts,he2025asap,peng2022ase}. In recent years, many studies have made significant progress on multiple tasks, including single-human motion imitation~\cite{luo2023perpetual,wang2025experts,he2025asap,tessler2024maskedmimic,liu2024mimicking}, human-object motion imitation~\cite{peng2018deepmimic,luo2023perpetual,peng2021amp,peng2022ase,tessler2024maskedmimic}, human-scene interaction~\cite{liu2024mimicking}, and multi-human interaction~\cite{takestwo2025zuhong}.

Humans can easily move through cluttered spaces. Even in scenes filled with furniture, narrow gaps, and irregular geometry, we naturally avoid collisions while maintaining the intent of our motion. Replicating this capability for physically simulated humanoids remains challenging, especially for the imitation of Human–Scene Interaction (HSI) within complex environments. This task requires tracking the given reference trajectory as faithfully as possible while respecting the geometric constraints imposed by complex scene meshes. However, the existing HSI imitation learning approach focuses only on overly simplified scenes and limited HSI behavior. For instance, MimicBench~\cite{liu2024mimicking} typically contains a single object as scene configurations and includes only six predefined tasks, such as sitting on a chair or lying on a sofa (\cref{fig:figure_0_motivation_illustration}).

In this paper, we focus on how to mimic human motions, captured from MoCap data, that dynamically interact with complex scenes. As illustrated in \cref{fig:figure_0_motivation_illustration}(a), a humanoid is expected to imitate reference motions while avoiding collisions with complex scene geometry. Through our investigation, we empirically observe an important pattern. As illustrated in \cref{fig:figure_0_motivation_illustration}(b), the early termination threshold $\tau$, which controls the maximum allowed deviation from the reference motion during training, has a significant impact on both the success rate and MPJPE of the learned policy. A larger $\tau$ improves task feasibility by allowing greater flexibility to avoid collisions, but degrades motion fidelity. Conversely, a smaller $\tau$ imposes stricter motion tracking but limits the policy’s ability to explore feasible motions, leading to premature failures in complex scenes.

Motivated by this challenge, we propose a two-stage method, \textbf{ComplexMimic}, to address the trade-off between motion fidelity and task feasibility. In stage I, we present the Dual Flow Strategy that learns two specialized expert policies. Specifically, we first train an \textit{imitation expert} in a scene-free setting using a strict termination threshold $\tau \le 0.25$, emphasizing precise motion tracking and high motion fidelity. Then, we learn another \textit{interaction expert} that leverages the full scene mesh and additional scene-aware information (\ie, height map). With a relaxed termination threshold, \ie, $\tau\ge0.5$, this expert enables robust adaptation to environmental geometry and physical collision handling.

In stage II, we distill the complementary expertise of the two experts into a single unified policy via multi-teacher knowledge distillation~\cite{jacobs1991adaptive,hinton2015distilling}. However, naive distillation strategies using uniform or random sampling overlook differences in expert learning difficulty, which may overtrain on easy samples, and consequently exhibit low efficiency. To tackle this problem, we introduce \textbf{D}ifficulty-\textbf{A}ware \textbf{D}istillation (DAD), which adaptively balances supervision across experts and prioritizes hard yet informative samples. We partition the training motions into two groups, \ie, imitation group and interaction group, so that each expert can specialize in the motions it handles best. The motion difficulty score is computed from online failure statistics, which guide inter-group sampling (favoring more difficult groups) and intra-group prioritization of harder samples.

We evaluate ComplexMimic on large-scale HSI datasets, including TRUMANS~\cite{jiang2024scaling} and LINGO~\cite{jiang2024autonomous}. Experimental results demonstrate that our approach achieves substantially better feasibility–faithfulness balance compared to existing baselines, especially achieving a \textbf{3.90\%} increase in Succ and a \textbf{9.36\%} reduction in $E_\text{g-mpjpe}$ on the TRUMANS dataset. Furthermore, zero-shot transfer to the real-world scanned dataset GIMO~\cite{zheng2022gimo} demonstrates strong robustness and generalization. Our contributions can be summarized as follows:

\begin{itemize}

\item  To the best of our knowledge, this is the first work systematically addressing human-scene interaction imitation under such realistic and complicated scene constraints. We propose a two-stage framework, ComplexMimic, to tackle this challenging problem.

\item We reveal a feasibility–faithfulness trade-off, where relaxing the early termination tracking constraint improves interaction feasibility but degrades motion fidelity, and vice versa.

\item We propose a Dual-Flow Strategy, which trains an imitation expert with strict tracking constraint for motion faithfulness and an interaction expert with relaxed tracking constraint and scene observations for feasible scene interaction.

\item We introduce Difficulty-Aware Distillation, a multi-teacher distillation strategy that adaptively balances supervision across experts and prioritizes hard-yet-learnable motion trajectories based on online motion-wise failure statistics and learning progress signals.

\end{itemize}

\section{Related Work}

\subsection{Humanoid Imitation Learning}
Physics-based imitation learning has become an efficient and practical paradigm for training humanoid controllers \cite{wang2025experts,wang2025skillmimic,xu2025intermimic,yu2025skillmimic,he2024omnih2o,liao2025beyondmimic,he2025asap,liu2024mimicking,xu2025learning}. Early work, such as DeepMimic ~\cite{peng2018deepmimic}, demonstrates that reinforcement learning guided by tailored imitation rewards can reproduce highly dynamic motions with stable physical control. Subsequent efforts relaxed strict motion alignment requirements. AMP~\cite{peng2021amp} introduces a novel discriminator-based objective, allowing more flexible matching between simulated motion and data. ASE~\cite{peng2022ase} further improves motion diversity by learning reusable low-level skills that can be composed for downstream tasks. More recently, approaches that focus on scaling control capacity and improving tracking robustness have been introduced. PHC~\cite{luo2023perpetual} adopts a Mixture-of-Experts (MoE) \cite{jacobs1991adaptive} paradigm and a negative sample mining strategy to enhance training efficiency. MaskedMimic~\cite{tessler2024maskedmimic} leverages a transformer-based full-body controller to achieve better tracking performance and utilizes a masked motion modeling to achieve more flexible humanoid robot control.

Although promising results have been made in humanoid imitation learning, these advances are mainly focused on simplified environments without scene constraints or with limited scene settings \cite{peng2018deepmimic,luo2023perpetual,tessler2024maskedmimic,liu2024mimicking}. In contrast, we target physics-based human–scene interaction imitation under dense and complicated 3D environments,
where geometric constraints introduce frequent collision and instability. We validate our framework on recent large-scale human–scene interaction datasets, including TRUMANS~\cite{jiang2024scaling} and LINGO~\cite{jiang2024autonomous}, which provide substantially more challenging and realistic interaction scenarios.

\subsection{Human-Scene Interaction}
Human-Scene Interaction (HSI) is a long-standing problem in human motion modeling \cite{zhang2023generating, guo2022generating,guo2022tm2t, zhang2025energymogen, zhang2024motiondiffuse,tevet2022human,zhang2026towards} and animation synthesis \cite{collorone2025monster,coins,hassan2019resolving,hassan2023synthesizing,hwang2025scenemi,zhao2023synthesizing,pan2024synthesizing,hassan2021stochastic,wang2024move,huang2022capturing,wang2025sims,shen2025detach,liu2024revisit}. A major line of work in HSI focuses on kinematic modeling, which models body-scene spatial relationships to generate motions consistent with scene geometry without requiring a closed-loop physics controller. PROX \cite{hassan2019resolving} is an early and influential work on this line, which proposes a novel human-scene interaction dataset based on RGB sequences in 12 scenes with 20 subjects. Building on PROX \cite{hassan2019resolving}, POSA \cite{hassan2021populating} introduces a body-centric contact-and-semantics representation over SMPL-X \cite{pavlakos2019expressive} vertices and is trained on PROX to extend to unseen scenes with plausible contact and semantic consistency. Recent advances focus on end-to-end generative models. HUMANISE \cite{wang2022humanise} and LINGO \cite{jiang2024autonomous} tend to generate 3D human motions in scenes based on text prompts through diffusion models \cite{sohl2015deep,ho2020denoising,song2020score,song2020denoising}, while TRUMANS \cite{jiang2024scaling} experiments with an autoregressive diffusion model conditioned on action labels. Another line of research adopts a closed-loop physics controller. UNIHSI \cite{xiao2023unified} leverages large language models and a chain of contact to achieve human-scene interaction control. TokenHSI \cite{pan2025tokenhsi} proposes a transformer-based unified controller for human-scene interaction control and skill learning.

Although prior works have achieved progress in physics-based human-scene interaction control, full-body imitation learning for HSI with complex 3D environments is less explored. In contrast, we target physics-based HSI imitation learning, where a humanoid must reproduce a given reference motion as faithfully as possible under scene constraints. And we explicitly address the fidelity-feasibility conflict by decoupling motion-faithful tracking and scene-aware interaction into complementary experts, and reconciling them through adaptive distillation with hierarchical difficulty-aware sampling.

\section{Methodology}
\subsubsection{Task Formulation.}
The goal of human-scene interaction (HSI) imitation learning is to learn a policy $\pi$ that produces a simulated human-scene interaction motion sequence $\{q_t\}_{t=1}^T$, which closely matches a ground-truth reference $\{\hat{q}_t\}_{t=1}^T$ derived from the MoCap data. Given the physical simulation of the humanoid and the 3D scene constraints, the policy $\pi$  should also compensate for missing or inaccurate details in the dataset. In particular, each simulated motion $q_t$  is defined as $q_t = \{\theta_t, p_t\}$, where $\theta_t \in \mathbb{R}^{J \times 3}$ and $p_t \in \mathbb{R}^{J \times 3}$ represent the rotations and  positions for $J$ joints, respectively. All simulation states have the corresponding ground-truth values, denoted by the hat symbol. 

\begin{figure}[t] 
    \centering 
    \includegraphics[width=1\textwidth, trim=3.5cm 1.5cm 3cm 3.5cm, clip]{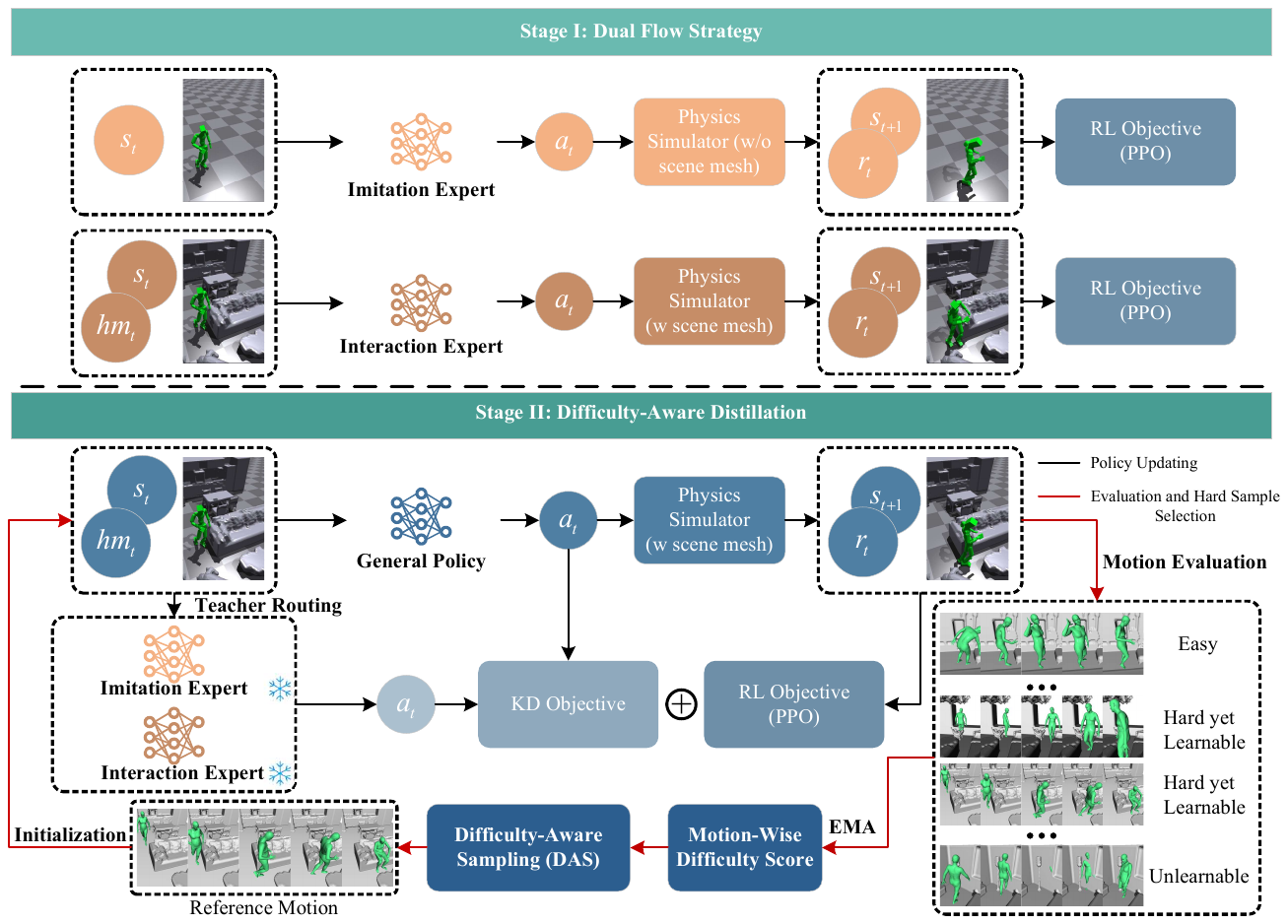}
    \caption{Overview of ComplexMimic, a two-stage framework for human–scene interaction imitation in complex environments. In Stage I, an imitation expert and an interaction expert are trained to achieve motion-faithful tracking and collision-aware adaptation, respectively. In Stage II, these experts are frozen and distilled into a unified policy via Difficulty-Aware Distillation (DAD). Black arrows denote policy updates; red arrows represent evaluation and hard-sample selection.}
    \label{fig:figure_1_pipeline} 
\end{figure}

\subsubsection{Overview.} 
We formulate HSI imitation as a Markov Decision Process (MDP) $(\mathcal{S}, \mathcal{A}, P, r, \gamma)$, defined by states $\mathcal{S}$, actions $\mathcal{A}$, simulator-provided transition dynamics $P$, reward function $r$, and discount factor $\gamma$. \cref{fig:figure_1_pipeline} illustrates our two-stage training framework. In Stage I, a Dual Flow Strategy trains decoupled teacher experts $\pi^{(T)}$: an imitation expert $\pi^{(T)}_\text{{Imitate}}$ for motion tracking accuracy and an interaction expert $\pi^{(T)}_\text{{Interact}}$ for interaction feasibility. In Stage II, we adaptively distill the knowledge of these teachers into a scalable student policy $\pi^{(S)}$ for whole-body control via Difficulty-Aware Distillation (DAD). \cref{Sec:Decoupled_Learning} details how teacher policies (\ie, experts) are trained with reinforcement learning, and \cref{sec:difficulty-aware-distillation} presents DAD for multi-teacher balancing and efficient mining of hard-yet-learnable motion samples.

\subsection{Dual Flow Strategy}

\label{Sec:Decoupled_Learning}
We propose a Dual Flow Strategy to train two teacher policies specialized for motion tracking and interaction, respectively.

\subsubsection{Imitation Expert.}
Given an initialized motion state $s_t$, the imitation expert $\pi^{(T)}_{\text{Imitate}}$ predicts an action $a_t$, which is executed in the physics simulator $\mathcal{E}(\mathcal{S})$ to obtain the next state $s_{t+1}$ and compute the imitation reward $r_t$. With a horizon length of $L_H$, the motion episode terminates when it satisfies an early termination condition or reaches the horizon length. After executing all motions in a minibatch, the policy is optimized using the PPO objective~\cite{schulman2017proximal}:
$\max_{\theta_I} \mathbb{E}\left[\sum_t \gamma^t r^{\text{imit}}(s_t, m_t)\right],$ where $r^{\text{imit}}$ denotes the imitation reward and $m_t$ represents the reference motion. Detailed definition of $r^{\text{imit}}$ is given in the supplementary material.

\subsubsection{Interaction Expert.}
\label{sec:scene_aware_expert}
The interaction expert $\pi^{(T)}_{\text{Interact}}(a_t|s_t, hm_t)$ is trained in the scene-constrained simulator $\mathcal{E}(\mathcal{S})$. It receives a height map~\cite{tessler2024maskedmimic} of the surrounding 3D environment $hm_t$ as additional scene perception. Specifically, the height map is represented by $20\times20$ samples on a square grid centered on the character and aligned with its heading direction, with neighboring samples spaced by $8\,\text{cm}$. The policy, therefore, conditions on both the humanoid state $s_t$ and the scene representation $hm_t$ to produce collision-aware actions while still tracking the reference motion using the imitation reward $r^{\text{imit}}$.

\subsubsection{Policy Learning.}
Following~\cite{peng2021amp}, the control policy $\pi$ is optimized using PPO~\cite{schulman2017proximal}. 
The policy gradient objective is defined as

\begin{equation}
\mathcal{L}_{\text{PPO}}(\theta)
=
\mathbb{E}_t \Big[
\min \big(
r_t(\theta) A_t,\,
\mathrm{clip}(r_t(\theta), 1-\epsilon, 1+\epsilon) A_t
\big)
\Big],
\end{equation}
where $\theta$ denotes the parameters of the policy $\pi$, and $r_t(\theta) =
\frac{\pi_\theta(a_t|s_t)}{\pi_{\theta_{\text{old}}}(a_t|s_t)}$ is the likelihood ratio between the updated and old policies. $\epsilon$ is a small clipping constant that constrains policy updates, and 
$A_t$ is the advantage estimate computed using the generalized advantage estimator (GAE$(\lambda)$) ~\cite{schulman2015high}.

\subsection{Difficulty-Aware Distillation}

\label{sec:difficulty-aware-distillation}

As illustrated in \cref{fig:figure_1_pipeline}, we utilize \textbf{D}ifficulty-\textbf{A}ware \textbf{D}istillation (DAD) to adaptively balance decoupled experts and prioritize hard yet learnable motions. Specifically, DAD consists of three components: (i) regime-specialized teacher routing, which assigns each motion to the most suitable expert, (ii) a multi-teacher distillation objective that transfers complementary knowledge to the student policy, and (iii) difficulty-aware sampling that prioritizes informative motions during training. Details of each component are given in this section.

\subsubsection{Regime-Specialized Teacher Routing.}
\label{sec:routing}

A naive multi-teacher distillation strategy typically combines supervision from multiple experts via action averaging or random routing, which often introduces gradient interference because different experts are optimized for fundamentally different objectives. As a result, the student may learn behaviors that are neither fully faithful to the reference motion nor sufficiently feasible under scene constraints.

To address this issue, we adopt regime-specialized teacher routing, which assigns each motion to the expert that handles it most reliably. We partition the training motions into two groups according to the tracking performance of the imitation expert in the scene-aware simulator: the tracking-friendly group $\mathcal{G}_{\mathrm{Imitate}}$ and the interaction-critical group $\mathcal{G}_{\mathrm{Inter}}$. Regime-specialized teacher routing is defined as
\begin{equation}
\pi^{(T)}(m)=
\begin{cases}
\pi^{(T)}_{\mathrm{Imitate}}, & m \in \mathcal{G}_{\mathrm{Imitate}},\\
\pi^{(T)}_{\mathrm{Interact}}, & m \in \mathcal{G}_{\mathrm{Inter}}.
\end{cases}
\end{equation}

The groups are constructed by evaluating $\pi^{(T)}_{\mathrm{Imitate}}$ on all training motions before distillation: successfully tracked motions are assigned to $\mathcal{G}_{\mathrm{Imitate}}$, and the rest to $\mathcal{G}_{\mathrm{Inter}}$. The effectiveness of regime-specialized routing is validated in \cref{sec:ablation}.

\subsubsection{Multi-Teacher Distillation Objective.}
\label{sec:kd}
Under the regime-specialized teacher routing, for each student transition $(s_t, a_t, r_t)$, we query the teacher actions
$a^{\text{imitate}}_t \sim \pi^{(T)}_{\mathrm{Imitate}}(\cdot \mid s_t)$
and $a^{\text{inter}}_t \sim \pi^{(T)}_{\mathrm{Interact}}(\cdot \mid s_t, hm_t)$.
Assuming Gaussian policies $\pi(a|s)=\mathcal{N}(\mu(s),\Sigma(s))$, distillation is implemented via the Kullback--Leibler divergence $D_{\mathrm{KL}}$ between action distributions:

\begin{equation}
\begin{aligned}
\mathcal{L}_{\text{KD}}(\theta)
=&\ \mathbb{I}\!\left[m\in\mathcal{G}_{\mathrm{Inter}}\right]\,
D_{\text{KL}}\!\left(\pi_\theta(\cdot|s_t,hm_t)\;\big\|\;\pi^{(T)}_{\mathrm{Interact}}(\cdot|s_t,hm_t)\right)\\
&+\Big(1-\mathbb{I}\!\left[m\in\mathcal{G}_{\mathrm{Inter}}\right]\Big)\,
D_{\text{KL}}\!\left(\pi_\theta(\cdot|s_t,hm_t)\;\big\|\;\pi^{(T)}_{\mathrm{Imitate}}(\cdot|s_t)\right).
\end{aligned}
\end{equation}

Distillation alone is insufficient because the student must interact with the environment under its own state distribution in $\mathcal{E}(\mathcal{S})$. Therefore, the student is trained with a joint objective:
\begin{equation}
\mathcal{L}(\theta) 
= 
\lambda_{\text{PPO}}\,\mathcal{L}_{\text{PPO}}(\theta) 
+ 
\lambda_{\text{KD}}\,\mathcal{L}_{\text{KD}}(\theta),
\end{equation}
where $\mathcal{L}_{\text{PPO}}$ denotes the standard clipped PPO objective computed from rollouts of $\pi_\theta$ in $\mathcal{E}(\mathcal{S})$, and $\mathcal{L}_{\text{KD}}$ is the distillation loss. 
The coefficients $\lambda_{\text{PPO}}$ and $\lambda_{\text{KD}}$ balance reinforcement learning and distillation.

\subsubsection{Difficulty-Aware Sampling.}

\label{sec:Difficulty-Aware Sampling}
Uniform or random sampling is a common choice in multi-teacher distillation with imitation learning\cite{wang2025experts, xu2025intermimic}, implicitly assuming that all motions contribute equally. In complex 3D scenes, this assumption breaks down: difficulty is highly skewed, and a small subset of motions accounts for most failures. Uniform sampling, therefore, wastes computation on already-solved trajectories while under-training feasibility-critical ones. To address this challenge, we propose the \textbf{D}ifficulty-\textbf{A}ware \textbf{S}ampling (DAS), which reallocates training budget based on the student’s online difficulty and learnability signals.

We first define the motion-wise difficulty score with online failure statistics. Let $m \in \mathcal{M}$ denote a motion clip. Every $K$ iterations, we evaluate the current student on motion $m$ and compute a binary failure indicator $f_k(m)=\mathbb{I}[\text{failure on } m \text{ at checkpoint } k]$, where ``failure'' is defined as violating the tracking-success criterion, namely when the average joint distance to the reference exceeds 0.5\,\text{m} at any timestep. We maintain a smoothed motion difficulty score via EMA as $D_k(m)\leftarrow(1-\beta)D_{k-1}(m)+\beta f_k(m)$, so $D_k(m)\in[0,1]$ measures how persistently motion $m$ fails under the current student.

Then, we utilize the motion-wise difficulty score to achieve inter-group sampling. To avoid collapsing into easy regimes, we perform inter-group sampling based on the motion difficulty score. We partition motions into aforementioned $C$ groups $\{\mathcal{G}_c\}_{c=1}^C$. The class difficulty is defined by aggregating motion difficulties:
\begin{equation}
D_k^{\mathrm{cls}}(c)=\frac{1}{|\mathcal{G}_c|}\sum_{m\in\mathcal{G}_c} D_k(m),
\label{eq:dad_class_diff}
\end{equation}
and we sample a group according to a softmax distribution:
\begin{equation}
p_k(c)=\frac{\exp(\tau_{\mathrm{inter}}\,D_k^{\mathrm{cls}}(c))}
{\sum_{c'=1}^{C}\exp(\tau_{\mathrm{inter}}\,D_k^{\mathrm{cls}}(c'))},
\label{eq:dad_cls_sampling}
\end{equation}
where $\tau_{\mathrm{inter}}$ controls how strongly we focus on difficult groups.

We further prioritize informative motion samples via intra-group prioritization. After selecting a group $c \sim p_k(c)$, we perform intra-group prioritization to mine the most informative motions within the group. We define a per-motion mining score $s_k(m)$ and sample motions using:
\begin{equation}
p_k(m \mid c)=\frac{\exp(\tau_{\mathrm{intra}}\, s_k(m))}
{\sum_{m'\in \mathcal{G}_c}\exp(\tau_{\mathrm{intra}}\, s_k(m'))},
\label{eq:dad_mot_sampling}
\end{equation}
A simple and effective choice is to set $s_k(m)$ proportional to the motion difficulty, $s_k(m)=D_k(m)$. Therefore, motions that fail more frequently are mined more aggressively within each group.

\begin{figure}[t] 
    \centering 
    \includegraphics[width=1\linewidth, trim=0.5cm 4cm 0.5cm 0.5cm, clip]{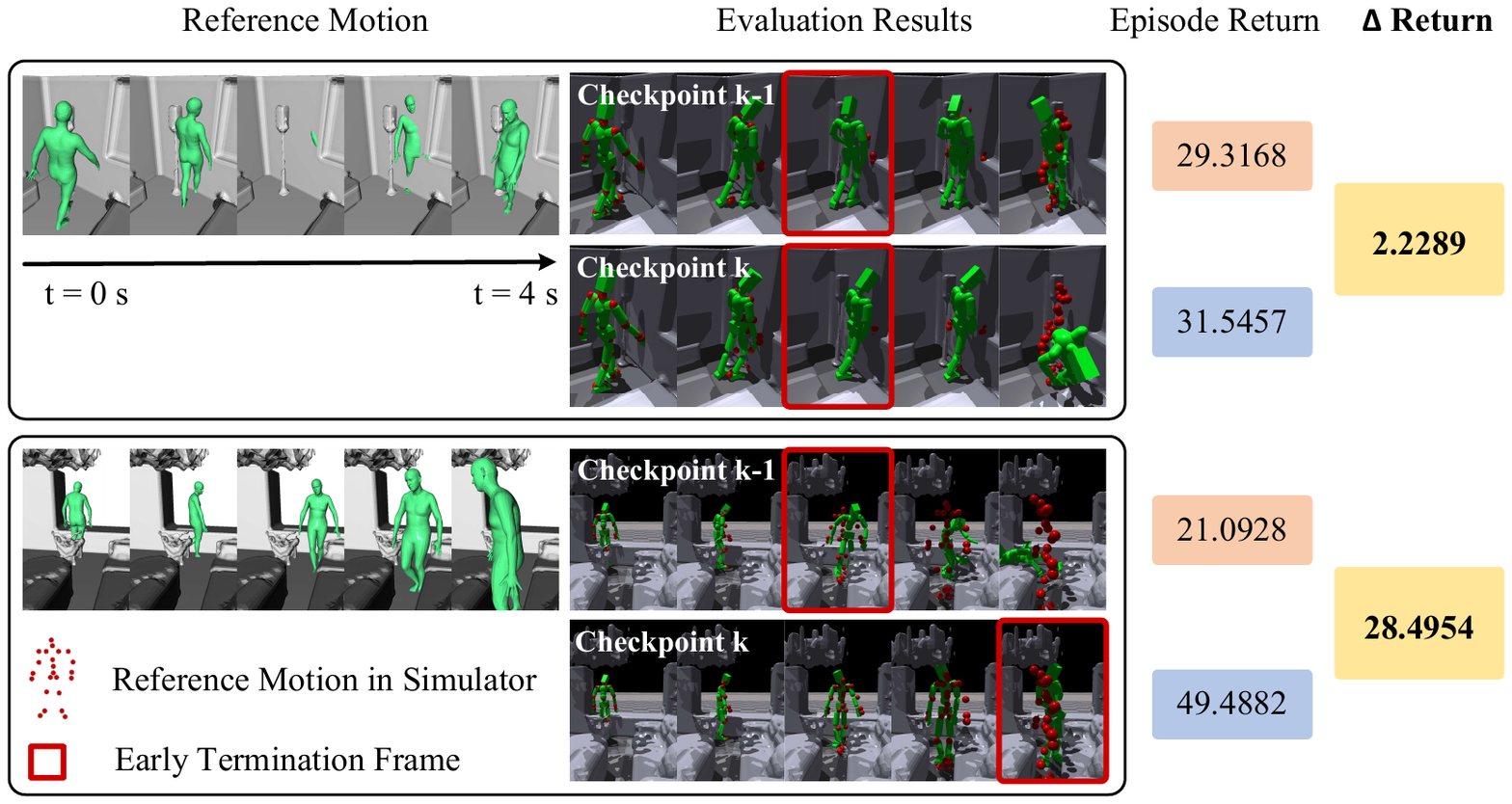}
    \caption{Examples of return improvement under different motion cases. The first reference motion conflicts heavily with the scene meshes, resulting in limited return gains. In contrast, the second motion achieves better tracking performance and larger return gains, demonstrating the effectiveness of our learning progress signal. }

    \label{fig:illustration_of_learnable_filter} 
\end{figure}

 Naïve intra-group hard mining may over-focus on motions that remain stagnant for long periods (e.g., highly disturbed or effectively infeasible cases), wasting computation and destabilizing training. We therefore introduce a learnability signal based on return improvement. Let $R_k(m)$ be the student’s episode return on motion $m$ at checkpoint $k$. We define the return improvement as $\Delta R_k(m)=R_k(m)-R_{k-1}(m)$, and maintain its exponential moving average (EMA) as $I_k(m)\leftarrow (1-\eta)I_{k-1}(m)+\eta\,\Delta R_k(m)$.

As shown in \cref{fig:illustration_of_learnable_filter}, some motions exhibit consistent return improvements, while others remain stagnant. If improvement is below a small threshold $\epsilon_u$, we downweight the sampling probability with:
\begin{equation}
u_k(m)=\exp\!\Big(-\lambda_u \,\max(0,\epsilon_u - I_k(m))\Big),
\label{eq:dad_unlearnable}
\end{equation}
where $\lambda_u$ controls the strength of penalization. We then incorporate this term into intra-group prioritization by redefining 
the mining score as $s_k(m)=D_k(m)+\log u_k(m)$. This keeps mining focused on motions that are hard yet learnable, while suppressing stagnant outliers.

 Putting everything together, the final probability of sampling motion $m$ is 
\begin{equation}
p_k(m)
=
(1-\epsilon)\, p_k(c(m))\, p_k(m\mid c(m))
+
\epsilon \cdot \frac{1}{|\mathcal{M}|},
\label{eq:dad_final}
\end{equation}
where $c(m)$ maps motion $m$ to its group, and 
$\epsilon$ controls the strength of a small uniform prior over motions.

\section{Experiments}

\subsection{Experiments Setup}
\label{sec:experiments_setup}

\subsubsection{Datasets.}
All the policies are trained on the TRUMANS dataset \cite{jiang2024scaling}, which is the largest high-quality human-scene interaction dataset with diverse and challenging scene geometries. Following SceneMI \cite{hwang2025scenemi}, motion sequences are segmented into 121 frames, with 10\% randomly selected for the test set. For evaluation, we tested our approach on three settings: (i) in-domain evaluation on TRUMANS \cite{jiang2024scaling} dataset; (ii) out-of-domain evaluation on unseen scene conditions on LINGO \cite{jiang2024autonomous} dataset; and (iii) real-world evaluation on GIMO \cite{zheng2022gimo} dataset, which consists of scanning real-world scene meshes with reconstruction artifacts and noises. 
\par

\subsubsection{Metrics.} 
Following PHC \cite{luo2023perpetual}, we utilize a series of pose-based and physics-based metrics to evaluate our motion imitation performance. We report the success rate (Succ), deeming imitation unsuccessful when, at any point during imitation, the body joints are on average $> 0.5\text{m}$ from the reference motion.  The global mean per-joint position error (MPJPE) $E_{\text{g-mpjpe}}$ and root-relative MPJPE $E_{\text{mpjpe}}$ are presented to measure the imitator's ability of imitating the reference motion both globally and locally (root-relative). To show physical realism, we also compare acceleration $E_{\text{acc}}$ ($\text{mm/frame}^2$) and velocity $E_{\text{vel}}$ ($\text{mm/frame}$) difference between simulated and MoCap motion. Detailed definitions of these metrics are provided in the supplementary material.

\subsubsection{Implementation Details.} 
All models are trained on a single NVIDIA 4090 GPU.  Physics simulation is carried out in NVIDIA's Isaac Gym~\cite{makoviychuk2021isaac}. The control policy is run at 30~Hz, while the simulation runs at 60~Hz. Following \cite{luo2023perpetual}, we do not consider body shape variation and use the mean SMPL body shape for evaluation. The temperatures for inter-group sampling and intra-group prioritization are both fixed to 1.0. Uniform mixture $\epsilon$ is set at a ratio $0.05$. EMA of Failure and return statistics are set to $0.95$ and $0.9$, respectively. The return improvement threshold $\epsilon_u$ is 0.1, and $\lambda_u$ is chosen as 1.0. Expert policies are implemented as MLPs with hidden sizes $\{2048, 1024, 1024, 512\}$. The student policy is implemented as a four-layer Transformer encoder with 4 heads, a hidden size of 512, and a feed-forward layer of 2048. The critic model is implemented as an MLP with hidden sizes $\{1024, 512\}$. We use $\tau=0.25$ and $\tau=0.5$ for the imitation expert and interaction expert, respectively. For student training, we keep $\tau$ fixed at $0.5$ to encourage physically feasible rollouts. Additional implementation details are provided in the supplementary material.

\par

\subsubsection{Baseline Models.}
We compare ComplexMimic with four state-of-the-art physics-based motion imitation methods: DeepMimic~\cite{peng2018deepmimic}, AMP~\cite{peng2021amp}, PHC~\cite{luo2023perpetual}, and MaskedMimic~\cite{tessler2024maskedmimic}. 
For fair comparison, we use the official implementations and augment all baselines with height maps at the same scale as ours.
\par

\begin{table}[t]
\centering
\caption{Quantitative comparisons on the TRUMANS \cite{jiang2024scaling} dataset. Best results are in \textbf{bold} and second-best are \underline{underlined}.}
\label{tab:trumans_evaluation}
\setlength{\tabcolsep}{6pt}
\renewcommand{\arraystretch}{1.15}
\begin{tabular}{l|ccccc}
\toprule
Method
& Succ $\uparrow$ & $E_{\text{g-mpjpe}}$ $\downarrow$ & $E_{\text{mpjpe}}$ $\downarrow$ & $E_{\text{acc}}$ $\downarrow$ & $E_{\text{vel}}$ $\downarrow$ \\
\midrule
DeepMimic~\cite{peng2018deepmimic}          & 0.674 & 99.002 & 73.349 & 16.335 & 14.823 \\
AMP~\cite{peng2021amp}          & 0.802 & 94.670 & 76.022  & 17.366 & 14.122 \\
PHC~\cite{luo2023perpetual}          & 0.853 & 85.793 & 65.365 & 13.559 & 12.055 \\
MaskedMimic~\cite{tessler2024maskedmimic} & 0.872 & 120.260 & 105.099 & 26.913 & 24.717 \\
\midrule
Ours w/o DAS                        & \underline{0.878} & \underline{84.026} & \underline{64.213} & \textbf{12.759} & \underline{11.194} \\
Ours                                & \textbf{0.906} & \textbf{77.840} & \textbf{64.128} & \underline{12.931} & \textbf{10.721} \\
\bottomrule
\end{tabular}

\end{table}

\begin{table}[t]
\centering
\caption{Quantitative comparisons on the LINGO \cite{jiang2024autonomous} dataset.}
\label{tab:lingo_evaluation}
\setlength{\tabcolsep}{6pt}
\renewcommand{\arraystretch}{1.15}
\begin{tabular}{l|ccccc}
\toprule
Method
& Succ $\uparrow$ & $E_{\text{g-mpjpe}}$ $\downarrow$ & $E_{\text{mpjpe}}$ $\downarrow$ & $E_{\text{acc}}$ $\downarrow$ & $E_{\text{vel}}$ $\downarrow$ \\
\midrule
DeepMimic~\cite{peng2018deepmimic}          & 0.535 & 128.974  & 92.421 & 24.079 & 20.907 \\
AMP~\cite{peng2021amp}          & 0.635 & 124.200 & 93.327 & 25.679 & 20.652 \\
PHC~\cite{luo2023perpetual}          & 0.615 & \textbf{120.472} & \textbf{84.782} & 23.648 & 20.383 \\
MaskedMimic~\cite{tessler2024maskedmimic} & \underline{0.719} & 163.525 & 132.315 & 38.387 & 36.499 \\
\midrule
Ours w/o DAS                        & 0.678 & \underline{122.046} & \underline{87.072} & 21.643 & \underline{18.595} \\
Ours & \textbf{0.746} & 124.665 & 90.575 & \textbf{20.641} & \textbf{17.303} \\
\bottomrule
\end{tabular}

\end{table}

\begin{table}[t]
\centering
\small
\setlength{\tabcolsep}{4pt}
\renewcommand{\arraystretch}{1.15}
\caption{Quantitative comparisons on the GIMO \cite{zheng2022gimo} dataset.}
\label{tab:gimo_evaluation}
\begin{tabular}{l|ccccc}
\toprule
Method
& Succ $\uparrow$ & $E_{\text{g-mpjpe}}$ $\downarrow$ & $E_{\text{mpjpe}}$ $\downarrow$ & $E_{\text{acc}}$ $\downarrow$ & $E_{\text{vel}}$ $\downarrow$ \\
\midrule
DeepMimic~\cite{peng2018deepmimic}          & 0.365 & 178.929 & 114.317 & 46.385 & 38.374 \\
AMP~\cite{peng2021amp}          & 0.446 & 173.632 & 113.682 & 46.974 & 37.353 \\
PHC \cite{luo2023perpetual}
& 0.443 & 173.580 & 109.696 & 47.151 & 38.556  \\
MaskedMimic \cite{tessler2024maskedmimic}
& 0.405 & 286.629 & 191.210 & 69.821 & 68.460  \\
\midrule
Ours w/o DAS & 0.515 & \underline{165.117} & \textbf{102.009} & \underline{38.738} & \underline{31.643} \\
Ours   & \textbf{0.579} & \textbf{161.969} & \underline{104.590} & \textbf{36.845} & \textbf{29.749} \\
\bottomrule
\end{tabular}

\end{table}

\subsection{Quantitative Evaluation}
\label{sec:quantitative_evaluation}
As shown in \cref{tab:trumans_evaluation}, our method consistently outperforms the competitive approaches on the TRUMANS~\cite{jiang2024scaling} test set across all metrics. In particular, Succ improves from 0.872 to 0.906 compared to the strongest baseline MaskedMimic~\cite{tessler2024maskedmimic}, while simultaneously reducing both $E_{\text{g-mpjpe}}$ and $E_{\text{mpjpe}}$. This demonstrates that ComplexMimic effectively resolves the feasibility–fidelity trade-off in complicated scene settings. \cref{tab:lingo_evaluation} presents out-of-domain evaluation results on the LINGO dataset~\cite{jiang2024autonomous} with unseen scene configurations. Our method achieves the highest Succ while maintaining competitive tracking accuracy compared to PHC~\cite{luo2023perpetual}, indicating strong cross-scene generalization. \cref{tab:gimo_evaluation} further reports real-world evaluation results on the GIMO dataset~\cite{zheng2022gimo}, which contains scanned environments with reconstruction noise. Our approach consistently surpasses the baseline methods across all metrics, achieving the best overall performance. These results highlight the robustness and generalization capability of our dual flow strategy and difficulty-aware distillation framework. We further analyze the contribution of individual components in \cref{sec:ablation}.

\begin{figure}[t] 
    \centering 
    \includegraphics[width=1\textwidth, trim=7cm 3.4cm 7cm 3cm, clip]{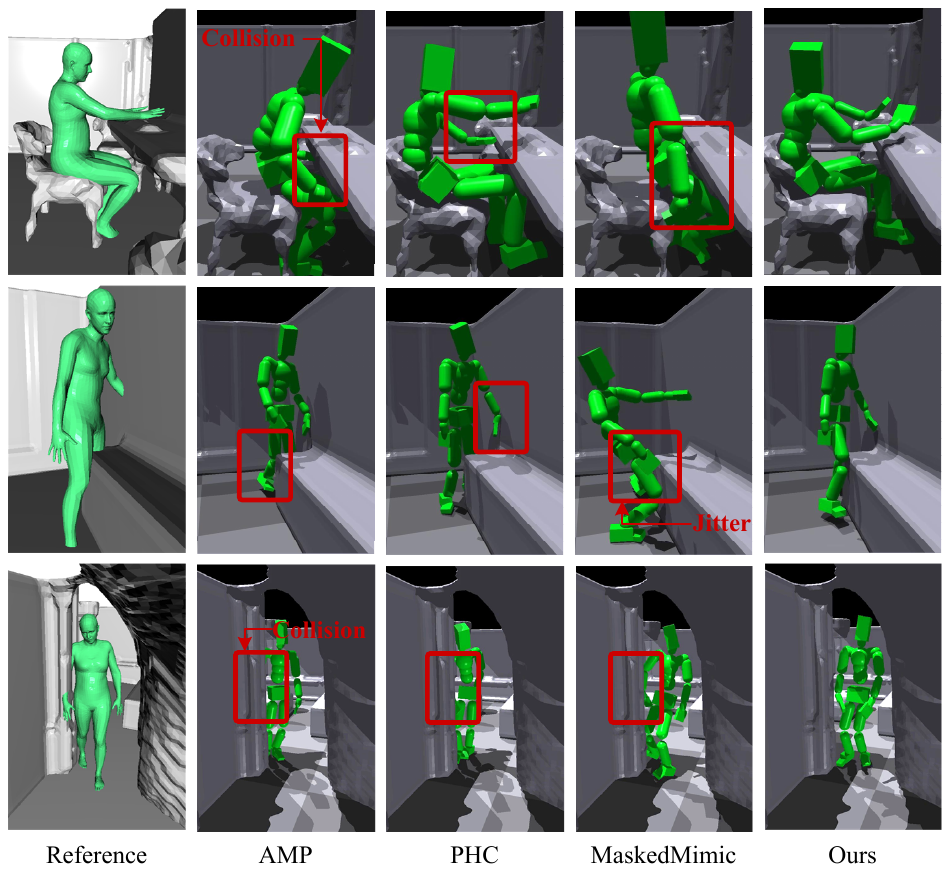}
    \caption{Qualitative comparisons on the TRUMANS \cite{jiang2024scaling} dataset. Red boxes highlight typical failure modes under complicated scenes, such as undesired collisions, unstable contacts, and noticeable deviations from the reference. More visual results can be found in the supplementary material.}

    \label{fig:qualitative_comparision_on_TRUMANS} 
\end{figure}

\subsection{Qualitative Evaluation}
\label{sec:qualitative_comparision}
To further demonstrate the advantage of our method against other competing baselines,
\cref{fig:qualitative_comparision_on_TRUMANS} presents visual examples on the TRUMANS \cite{jiang2024scaling} test set. AMP~\cite{peng2021amp} often fails to robustly track the reference in complicated scenes, leading to noticeable penetrations and unstable contacts. While PHC~\cite{luo2023perpetual} can accurately track the reference motion, it is prone to being disturbed by the 3D scene meshes. As for the MaskedMimic~\cite{tessler2024maskedmimic}, although it can achieve feasible HSI tracking, the deviations from the reference are rather obvious. In contrast, our approach produces more feasible interactions by correcting incorrect body configurations while preserving the overall motion structure. 

\begin{figure}[t] 
    \centering 
    \includegraphics[width=1\textwidth, trim=5cm 4.1cm 8cm 6cm, clip]{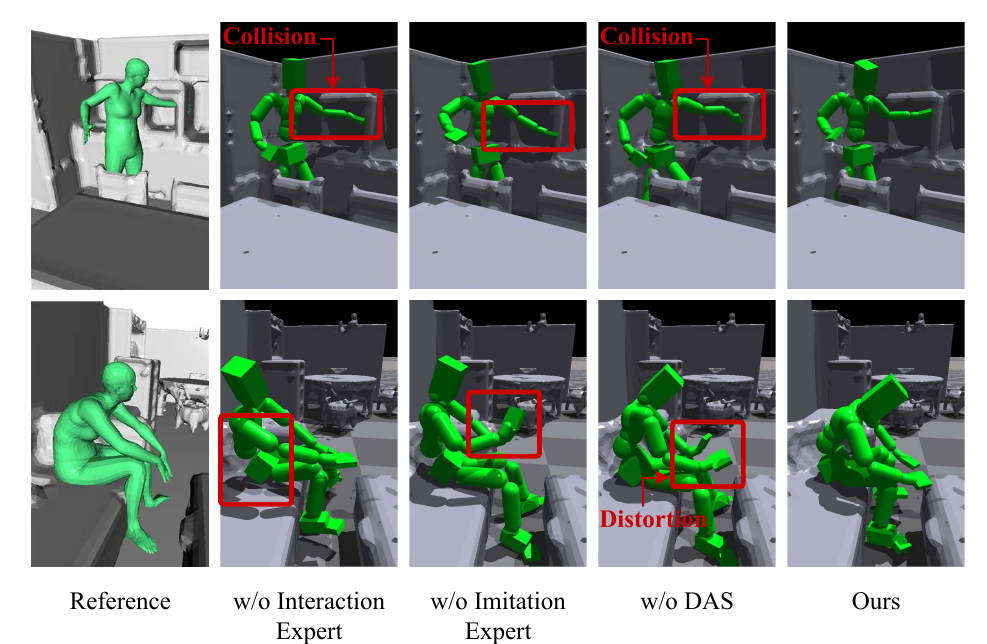}
    \caption{ Qualitative ablation comparisons on TRUMANS~\cite{jiang2024scaling}.}
 
    \label{fig:ablation_qualitative_comparision} 
\end{figure}

\begin{table}[t]
\centering
\small
\setlength{\tabcolsep}{4pt}
\renewcommand{\arraystretch}{1.15}
\caption{Ablation study on TRUMANS \cite{jiang2024scaling}. “w/o” indicates removal; “–” indicates replacement of regime-specialized teacher routing. Additional ablation studies on LINGO~\cite{jiang2024autonomous} and GIMO~\cite{zheng2022gimo} are provided in the supplementary material.}
\label{tab:ablation_TRUMANS}
\begin{tabular}{l|ccccc}
\toprule
Method
& Succ $\uparrow$ & $E_{\text{g-mpjpe}}$ $\downarrow$ & $E_{\text{mpjpe}}$ $\downarrow$ & $E_{\text{acc}}$ $\downarrow$ & $E_{\text{vel}}$ $\downarrow$ \\
\midrule
w/o Imitation Expert 
& 0.876 & 86.059 & 64.487 & 13.170 & 11.477 \\

w/o Interaction Expert
& 0.847 & 82.489 & \underline{62.464} & \textbf{12.565} & 11.416 \\

w/o Inter-Group Sampling 
& \underline{0.899} & 82.045 & 62.816 & \underline{12.779} & \underline{10.918} \\

w/o Intra-Group Prioritization 
& 0.898 & \underline{80.735} & \textbf{61.016} & 13.349 & 11.248 \\

w/o Learnability Filtering 
& 0.877 & 88.497 & 67.465 & 14.231 & 11.876 \\

w/o PPO
& 0.830 & 110.880 & 79.460 & 18.788 & 16.475 \\

- Random Routing 
& 0.874 & 83.016 & 63.715 & 12.625 & 10.945 \\

- Action Averaging 
& 0.873 & 83.698 & 64.075 & 12.712 & 11.030 \\

\midrule
\textbf{Ours} 
& \textbf{0.906} & \textbf{77.840} & 64.128 & 12.931 & \textbf{10.721} \\
\bottomrule
\end{tabular}

\end{table}

\subsection{Ablation Study}
\label{sec:ablation}

\subsubsection{Effectiveness of Dual Flow Strategy.}
As shown in \cref{tab:ablation_TRUMANS}, removing either teacher degrades the performance significantly. Without the imitation expert, tracking error $E_\text{g-mpjpe}$ degrades (from 77.840 to 86.059), while removing the interaction expert leads to a notable drop in Succ (from 0.906 to 0.847), highlighting the importance of interaction-aware supervision. Qualitative results in \cref{fig:ablation_qualitative_comparision} further confirm that the two experts provide complementary guidance for fidelity and feasibility. 

\subsubsection{Effectiveness of Difficulty-Aware Sampling (DAS).}
Each component of DAS consistently contributes to the final performance. Disabling intra-group prioritization or inter-group sampling reduces Succ and worsens tracking accuracy, showing that (i) focusing updates on harder motions within each regime and (ii) balancing training across regimes are both beneficial. The learnability filtering is particularly important: removing it significantly degrades both Succ (from 0.906 to 0.877) and $E_{\text{g-mpjpe}}$ (from 77.840 to 88.497), suggesting that suppressing hard but unlearnable clips is necessary for stable and efficient distillation.

\subsubsection{Effectiveness of joint optimization with PPO.}
As exhibited in \cref{tab:ablation_TRUMANS}, adding the PPO objective significantly improves both Succ (from 0.830 to 0.906) and  $E_{\text{g-mpjpe}}$ (from 110.880 to 77.840), highlighting the necessity of on-policy refinement beyond pure distillation.

\subsubsection{Effectiveness of regime-specialized teacher routing.}
To evaluate the necessity of regime-specialized teacher routing, we compare with two alternatives: (i) Random routing, where each motion is supervised by either expert with equal probability, and (ii) Action averaging, where the student imitates the mean action of both experts. As shown in \cref{tab:ablation_TRUMANS}, both variants degrade performance. Random routing introduces inconsistent supervision across motion regimes, while action averaging blends conflicting expert objectives, leading to gradient interference and suboptimal performance.

\section{Conclusion}
In this paper, we address an important yet underexplored problem: human-scene interaction imitation learning under complex scene constraints. To solve the scene-induced infeasibility and achieve accurate motion imitation, we propose a decoupled training paradigm, which leverages the imitation expert to track the reference motion faithfully and utilizes the interaction expert to interact with the 3D environment appropriately. To combine these complementary capabilities into a unified policy, we introduce difficulty-aware distillation, which adaptively balances supervision across different expert regimes and prioritizes hard-yet-learnable motions using failure statistics and return improvement signals. Extensive experiments on three human-scene interaction datasets show that our approach consistently improves both interaction feasibility and motion accuracy, and validate the contribution of each component.

\bibliographystyle{splncs04}
\bibliography{main}
\clearpage

\appendix
\begin{center}
{\Large \bfseries Supplementary Material \par}
\end{center}

In this supplementary material, we present:
\begin{itemize}

\item \cref{sec:ablation_on_lingo_and_gimo}: Additional ablation results on LINGO~\cite{jiang2024autonomous} and GIMO~\cite{zheng2022gimo}.
\item \cref{sec:sampling_temperatures}: Ablations on sampling temperatures for difficulty-aware sampling.
\item 
\cref{sec:learnability_threshold}: Ablations on the threshold of learnability filtering.
\item \cref{sec:details_of_Policy_Representation}: Details of policy representation.
\item \cref{sec:details_of_reward_function}: Details of reward function.
\item \cref{sec:details_of_evaluation_metrics}: Details of evaluation metrics.
\item \cref{sec:additinal_implementation_details}: Additional implementation details.
\item \cref{sec:cross_simulator_generalization}: Cross-simulator generalization results in Isaac Lab.
\item \cref{sec:cross_embodiment_generalization}: Cross-embodiment generalization results.
\item \cref{sec:discussion}: Discussion.

\end{itemize}

\section{Ablation Results on LINGO and GIMO}
\label{sec:ablation_on_lingo_and_gimo}

\cref{tab:LINGO_ablation} and \cref{tab:GIMO_ablation} report additional ablation results on LINGO~\cite{jiang2024autonomous} and GIMO~\cite{zheng2022gimo}, respectively. Removing any component consistently degrades performance on both datasets. In particular, removing either expert in the dual-flow strategy leads to a noticeable drop in both Succ and tracking accuracy, indicating that both experts provide complementary supervision signals. Disabling the regime-specialized teacher routing weakens the model's ability to select suitable supervision. Similarly, removing inter-group sampling or intra-group prioritization results in less effective training samples, which negatively affects both motion fidelity and interaction feasibility. In addition, removing PPO joint optimization during distillation leads to the largest performance drop (Succ from 0.746 to 0.541 on LINGO). These results collectively demonstrate that each component plays an important role in the final performance of the proposed framework.

\begin{table}[h]
\centering
\small
\setlength{\tabcolsep}{4pt}
\renewcommand{\arraystretch}{1.15}
\caption{Ablation study on the LINGO \cite{jiang2024autonomous}. “w/o” indicates removal; “–” indicates replacing teacher routing. Best results are in \textbf{bold} and second-best are \underline{underlined}.}
\label{tab:ablation_LINGO}
\begin{tabular}{l|ccccc}
\toprule
Method
& Succ $\uparrow$ & $E_{\text{g-mpjpe}}$ $\downarrow$ & $E_{\text{mpjpe}}$ $\downarrow$ & $E_{\text{acc}}$ $\downarrow$ & $E_{\text{vel}}$ $\downarrow$ \\
\midrule
w/o Imitation Expert 
& \underline{0.729} & 133.104 & 95.958 & \textbf{20.310} & 17.772 \\

w/o Interaction Expert 
& 0.701 & \underline{129.079} & 94.020 & 20.721 & \underline{17.562} \\

w/o Inter-Group Sampling
& 0.683 & 128.558 & \underline{91.545} & 20.593 & 17.868 \\

w/o Intra-Group Prioritization
& 0.678 & 134.071 & 98.466 & 20.772 & 18.024 \\

w/o Learnability Filtering 
& 0.653 & 132.283 & 97.877 & \underline{20.475} & 18.065 \\

w/o PPO
& 0.541 & 131.683 & 95.516 & 21.871 & 19.389 \\

- Random Routing 
& 0.654 & 132.350 & 98.218 & 20.527 & 18.044 \\

- Weighted Averaging 
& 0.633 & 129.875 & 95.274 & 21.802 & 18.973 \\

\midrule
Ours & \textbf{0.746} & \textbf{124.665} & \textbf{90.575} & 20.641 & \textbf{17.303} \\
\bottomrule
\end{tabular}
\label{tab:LINGO_ablation}
\end{table}

\begin{table}[h]
\centering
\small
\setlength{\tabcolsep}{4pt}
\renewcommand{\arraystretch}{1.15}
\caption{Ablation study on the GIMO \cite{zheng2022gimo}.}
\label{tab:ablation_GIMO}
\begin{tabular}{l|ccccc}
\toprule
Method
& Succ $\uparrow$ & $E_{\text{g-mpjpe}}$ $\downarrow$ & $E_{\text{mpjpe}}$ $\downarrow$ & $E_{\text{acc}}$ $\downarrow$ & $E_{\text{vel}}$ $\downarrow$ \\
\midrule
w/o Imitation Expert 
& 0.494 & 162.737 & 105.918 & 40.447 & 33.495 \\

w/o Interaction Expert
& 0.501 & 162.279 & 103.798 & 41.884 & 34.019 \\

w/o Inter-Group Sampling
& \underline{0.535} & 166.505 & 107.671 & 39.284 & \underline{31.404} \\

w/o Intra-Group Prioritization 
& 0.512 & \underline{162.294} & \textbf{100.672} & 39.471 & 32.610 \\

w/o Learnability Filtering 
& 0.493 & 162.374 & \underline{102.429} & 40.920 & 33.847 \\

w/o PPO
& 0.483 & 167.744 & 103.779 & 40.544 & 32.236 \\

- Random Routing 
& 0.520 & 164.078 & 106.842 & 38.851 & 31.614 \\

- Weighted Averaging 
& 0.531 & 164.332 & 106.956 & \underline{38.383} & 31.346 \\

\midrule
Ours   & \textbf{0.579} & \textbf{161.969} & 104.590 & \textbf{36.845} & \textbf{29.749} \\
\bottomrule
\end{tabular}
\label{tab:GIMO_ablation}
\end{table}

\section{Impact of Sampling Temperatures}
\label{sec:sampling_temperatures}

 We show the effect of sampling temperatures in \cref{tab:temperature_ablation}. We find that moderate temperatures provide the best balance between success rate and motion fidelity, with $\tau_{\text{inter}}=1.0$ and $\tau_{\text{intra}}=1.0$ yielding the best overall performance.

\begin{table}[h]
\centering
\caption{Ablations on sampling temperatures $\tau_{\text{inter}}$ and $\tau_{\text{intra}}$.}
\begin{tabular}{c c | c c c c c}
\toprule
$\tau_{\text{inter}}$ & $\tau_{\text{intra}}$ 
& Succ $\uparrow$ & $E_{\text{g-mpjpe}}$ $\downarrow$ & $E_{\text{mpjpe}}$ $\downarrow$ & $E_{\text{acc}}$ $\downarrow$ & $E_{\text{vel}}$ $\downarrow$ \\
\midrule
0.5 & 0.5 & 0.867 & 99.978 & 75.697 & 13.430 & 12.216 \\
0.5 & 1.0 & 0.869 & 99.574 & 75.691 & 13.488 & 12.277 \\
0.5 & 1.5 & 0.886 & 84.487 & 64.222 & \underline{12.786} & \underline{11.268} \\
\midrule
1.0 & 0.5 & \textbf{0.906} & 91.101 & 68.881 & 13.029 & 11.311 \\
1.0 & 1.0 & \textbf{0.906} & \textbf{77.840} & \underline{64.128} & 12.931 & \textbf{10.721} \\
1.0 & 1.5 & \underline{0.905} & 86.430 & 65.699 & 13.256 & 11.287 \\
\midrule
1.5 & 0.5 & 0.893 & 83.859 & \textbf{64.049} & \textbf{12.549} & 11.077 \\
1.5 & 1.0 & 0.901 & \underline{82.854} & 64.805 & 12.847 & 11.700 \\
1.5 & 1.5 & 0.898 & 86.507 & 65.841 & 13.063 & 11.299 \\
\bottomrule
\end{tabular}

\label{tab:temperature_ablation}
\end{table}

\section{Impact of Learnability Threshold}
\label{sec:learnability_threshold}
We investigate the effect of the learnability threshold $\epsilon_u$, and the results are reported in \cref{tab:epsilon_u_ablation}. Increasing the threshold filters out more corrupted reference motions and progressively improves performance. However, an overly large threshold leads to decreases in both Succ and $E_\mathrm{g\text{-}mpjpe}$, as many hard samples are removed during training. Across all tested values, the performance consistently outperforms the baseline with $\epsilon_u=0$, validating the robustness of the learnability filtering.

\begin{table}[t]
\centering
\caption{Ablations on threshold $\epsilon_u$ for learnability filtering.}

\begin{tabular}{c|ccccc}
\toprule
$\epsilon_u$ & Succ $\uparrow$ & $E_{\text{g-mpjpe}}$ $\downarrow$ & $E_{\text{mpjpe}}$ $\downarrow$ & $E_{\text{acc}}$ $\downarrow$ & $E_{\text{vel}}$ $\downarrow$ \\
\midrule
0     & 0.877 & 88.497 & 67.465 & 14.231 & 11.876 \\
0.05  & 0.892 & 86.008 & 66.154 & \textbf{12.378} & \underline{11.378} \\
0.1   & \textbf{0.906} & \textbf{77.84} & \textbf{64.128} & \underline{12.931} & \textbf{10.721} \\
0.2   & \underline{0.899} & \underline{81.136} & \underline{64.297} & 14.176 & 11.570 \\
\bottomrule
\end{tabular}
\label{tab:epsilon_u_ablation}
\end{table}

\section{Details of Policy Representation}
\label{sec:details_of_Policy_Representation}

We model HSI imitation as an MDP $(\mathcal{S}, \mathcal{A}, P, r, \gamma)$.
At each timestep $t$, the policy observes a state $s_t \in \mathcal{S}$ and outputs an action $a_t \in \mathcal{A}$. The simulator executes $a_t$ under physics-based transition dynamics $P$, and returns a reward $r_t = r(s_t,a_t)$ used for RL training. $\gamma$ denotes the discount factor.

\noindent\textbf{Action space.}
 Whole-body PD target control is used, where the action 
$a_t \in \mathbb{R}^{D}$ specifies target joint rotations for all controllable DoFs. Given $a_t$, the simulator computes joint torques using PD controllers.

\noindent\textbf{State space.}
The observation $s_t$ combines two groups of features:
\begin{equation}
s_t = \big[s_t^{\mathrm{prop}};\; s_t^{\mathrm{ref}}],
\end{equation}
where $[\cdot;\cdot]$ denotes the concatenation operator.

\noindent\textbf{Proprioceptive features $s_t^{\mathrm{prop}}$.}
Humanoid joint states
$\{\theta_t^{h},\, p_t^{h},\, \omega_t^{h},\, v_t^{h}\}$ are included in the proprioceptive features. $\theta_t^{h}$, $p_t^{h}$, $\omega_t^{h}$, and $v_t^{h}$ denote the joint rotation, position, angular velocities, and linear velocities, respectively.

\noindent\textbf{Reference-tracking features $s_t^{\mathrm{ref}}$.}
To perform imitation, phase-aligned references from motion clip $m$ are provided over a short future horizon $\{t+k\}_{k=1}^{H}$:
$\{\hat{\theta}^{h}_{t+k},\, \hat{p}^{h}_{t+k},\, \hat{\omega}^{h}_{t+k},\, \hat{v}^{h}_{t+k}\}_{k=1}^{H}$. Following \cite{luo2023perpetual}, $H$ is set to 1. Current tracking errors between simulation and reference are also provided, which is
defined as $\Delta x_t = \hat{x}_{t+1} - {x}_t$ for each root quantity. The full $s_t^{\mathrm{ref}}$ can be represented as:
\begin{equation}
s_t^{\mathrm{ref}}
=
\Big[
\{\hat{\theta}^{h}_{t+k},\, \hat{p}^{h}_{t+k},\, \hat{\omega}^{h}_{t+k},\, \hat{v}^{h}_{t+k}\}_{k=1}^{H};\ 
\Delta\theta_t^{h},\, \Delta p_t^{h},\, \Delta\omega_t^{h},\, \Delta v_t^{h}
\Big],
\end{equation}
where $\Delta\theta_t^{h}=\hat{\theta}_{t+1}^{h}-\theta_t^{h}$,
$\Delta p_t^{h}=\hat{p}_{t+1}^{h}-p_t^{h}$,
$\Delta\omega_t^{h}=\hat{\omega}_{t+1}^{h}-\omega_t^{h}$, and
$\Delta v_t^{h}=\hat{v}_{t+1}^{h}-v_t^{h}$.

\section{Details of Reward Function}
\label{sec:details_of_reward_function}

 Both the teacher and the student policies are trained using the same imitation reward, which is defined as:

\begin{equation}
r_t^{\text{imit}} =
0.5\,e^{-100\|\hat p_t-p_t\|}
+
0.3\,e^{-10\|\hat \theta_t \ominus \theta_t\|}
+
0.1\,e^{-0.1\|\hat v_t-v_t\|}
+
0.1\,e^{-0.1\|\hat \omega_t-\omega_t\|}.
\end{equation}

\section{Details of Evaluation Metrics}
\label{sec:details_of_evaluation_metrics}

\noindent\textbf{Success Rate (Succ).}
Succ measures whether the policy can successfully imitate the reference motion without failure. Formally, Succ of a motion clip $m$ is defined as:

\begin{equation}
\text{Succ}(m) =
\begin{cases}
1, & \text{if } \max\limits_{t} \frac{1}{J}\sum_{j=1}^{J} 
\left\| p_{t,j} - \hat{p}_{t,j} \right\|_2 < \delta,\\
0, & \text{otherwise},
\end{cases}
\end{equation}
where $J$ is the number of joints, $p_{t,j}$ and $\hat{p}_{t,j}$ denote the predicted and reference joint positions at timestep $t$, respectively. The final success rate is computed as the average over all motion clips.
Following \cite{luo2023perpetual}, $\delta$ is set to 0.5 m.

\noindent\textbf{Global Mean Per-Joint Position Error ($E_{\text{g-mpjpe}}$).}
$E_{\text{g-mpjpe}}$ measures the average euclidean distance between predicted and reference joint positions in the global coordinate system:

\begin{equation}
E_{\text{g-mpjpe}}
=
\frac{1}{T J}
\sum_{t=1}^{T}
\sum_{j=1}^{J}
\left\|
p_{t,j} - \hat{p}_{t,j}
\right\|_2,
\end{equation}
where $T$ denotes the length of the motion sequence.

\noindent\textbf{Root-relative Mean
Per-Joint Position Error ($E_{\text{mpjpe}}$).}
To evaluate local motion accuracy independent of global transition, we compute $E_{\text{mpjpe}}$ after removing the root joint position:

\begin{equation}
E_{\text{mpjpe}}
=
\frac{1}{T J}
\sum_{t=1}^{T}
\sum_{j=1}^{J}
\left\|
(p_{t,j} - p_{t,\text{root}})
-
(\hat{p}_{t,j} - \hat{p}_{t,\text{root}})
\right\|_2 .
\end{equation}

\noindent\textbf{Acceleration Error ($E_{\text{acc}}$).}
Acceleration error measures the difference between predicted and reference joint accelerations:

\begin{equation}
E_{\text{acc}}
=
\frac{1}{(T-2)J}
\sum_{t=2}^{T-1}
\sum_{j=1}^{J}
\left\|
a_{t,j} - \hat{a}_{t,j}
\right\|_2 ,
\end{equation}
where the joint acceleration is approximated by: $a_{t,j} = p_{t+1,j} - 2p_{t,j} + p_{t-1,j}.$

\noindent\textbf{Velocity Error ($E_{\text{vel}}$).}
Velocity error evaluates the difference between predicted and reference joint velocities:

\begin{equation}
E_{\text{vel}}
=
\frac{1}{(T-1)J}
\sum_{t=1}^{T-1}
\sum_{j=1}^{J}
\left\|
v_{t,j} - \hat{v}_{t,j}
\right\|_2 ,
\end{equation}
where the velocity is computed as: $v_{t,j} = p_{t+1,j} - p_{t,j}$.

\section{Additional Implementation Details}
\label{sec:additinal_implementation_details}

We import the scene meshes into Isaac Gym~\cite{makoviychuk2021isaac} following~\cite{xiao2023unified}. We also attempted to convert the meshes into simulation assets via convex decomposition, as used in~\cite{xu2025intermimic}. However, convex decomposition is time-consuming and significantly reduces the geometric fidelity. The hyper parameters used during PPO optimization are illustrated in \cref{tab:hyper_parameters_of_policies}. To evaluate the out-of-domain performance of our method, we select motion clips longer than 121 frames from LINGO~\cite{jiang2024autonomous}, resulting in 8,220 test motions. To evaluate performance in real-world scanned environments, we use the full motions from GIMO as the test set. Following~\cite{zheng2022gimo}, the motions are segmented into 121-frame clips, resulting in 561 motion clips in total. 

\begin{table}[h]
\centering
\caption{Hyper parameters for PPO optimization.}
\label{tab:hyper_parameters_of_policies}
\setlength{\tabcolsep}{8pt}
\renewcommand{\arraystretch}{1.1}
\begin{tabular}{l l}
\toprule
\textbf{Hyper parameters} & \textbf{Value} \\
\midrule
Action distribution & 69D Continuous \\
Discount factor $\gamma$ & 0.99 \\
Generalized advantage estimation $\lambda$ & 0.95 \\
Entropy regularization coefficient & 0.0 \\
Optimizer & Adam~\cite{kingma2014adam} \\
Learning rate (Actor-MLP) & $2\times10^{-5}$ \\
Learning rate (Actor-Transformer) & $2\times10^{-6}$ \\
Learning rate (Critic) & $1\times10^{-4}$ \\
Batch Size  & 512 \\
Minibatch size & 16384 \\
Horizon length $H$ & 32 \\
Action bounds loss coefficient & 10 \\
Maximum episode length & 300 \\
\bottomrule
\end{tabular}
\end{table}

\section{Cross-Simulator Generalization}
\label{sec:cross_simulator_generalization}

\begin{figure}[h]
\centering
\includegraphics[width=1\linewidth]{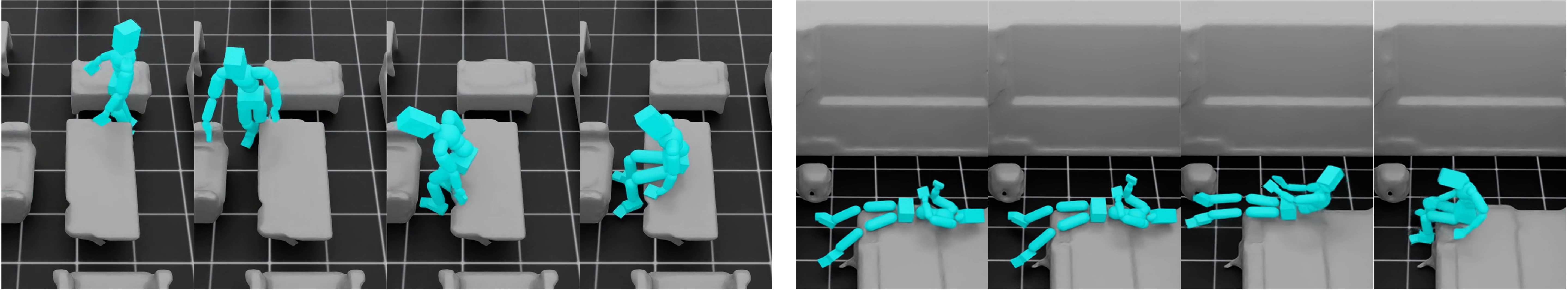}
{\caption{Visualization of our method in Isaac Lab.}
\label{fig:isaaclab_vis}
}
\end{figure}

To evaluate generalization beyond Isaac Gym, we conduct cross-simulator experiments by directly testing our method in Isaac Lab without fine-tuning. As shown in \cref{fig:isaaclab_vis}, our policy produces stable rollouts and plausible scene interactions under Isaac Lab dynamics. \cref{tab:cross_sim_generalization} further shows that our method outperforms the baselines.

\begin{table}[h]
\centering
{\caption{
Quantitative results of cross-simulator generalization.
}
\begin{tabular}{l|ccccc}
\toprule
Method 
& Succ. $\uparrow$ 
& $E_{\text{g-mpjpe}} \downarrow$ 
& $E_{\text{mpjpe}} \downarrow$ 
& $E_{\text{acc}} \downarrow$ 
& $E_{\text{vel}} \downarrow$ \\
\midrule
PHC~\cite{luo2023perpetual}          
& 0.612 &  235.475 & 142.545 & 8.764 &  12.918 \\
MaskedMimic~\cite{tessler2024maskedmimic}
& 0.548 & 272.741 & 165.120 & 8.681 & 13.742 \\
Ours         
&  \textbf{0.769} & \textbf{167.087} & \textbf{104.407} & \textbf{8.076} & \textbf{10.153} \\
\bottomrule
\end{tabular}

\label{tab:cross_sim_generalization}
}
\end{table}

\FloatBarrier

\section{Cross-Embodiment Generalization}
\label{sec:cross_embodiment_generalization}
\begin{figure}[h]
\centering
\includegraphics[width=1\linewidth]{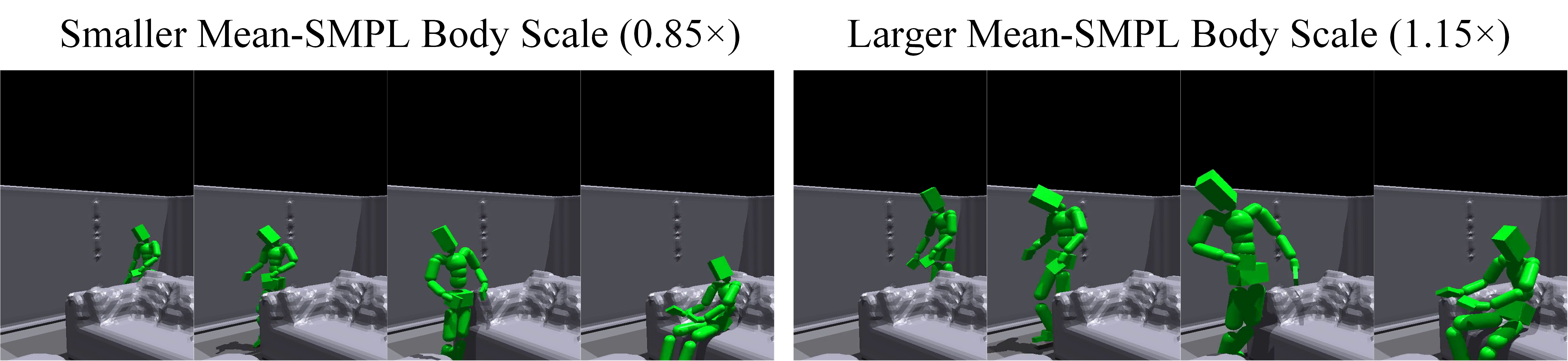}
{
\caption{Visualization under different body scales.}
\label{fig:shape_vis}
}
\end{figure}

  To evaluate cross-embodiment generalization, we directly test our policy on scaled mean-SMPL humanoids ($0.85\times$/$1.15\times$) without fine-tuning.  As visualized in \Cref{fig:shape_vis}, our method can stably roll out under different embodiments. \Cref{tab:cross_embodiment_generalization} further shows that our method is more robust than baselines under different scale settings, suggesting that it is not limited to a single fixed body scale.

\begin{table}[h]
{
\caption{Quantitative results under different body scales.}
\label{tab:cross_embodiment_generalization}
}
\begin{tabular}{
@{}l@{\hspace{5pt}}|
@{\hspace{2pt}}c@{\hspace{3pt}}c@{\hspace{3pt}}c@{\hspace{2pt}}|
@{\hspace{2pt}}c@{\hspace{3pt}}c@{\hspace{3pt}}c@{}}
\toprule
Method 
& \multicolumn{3}{c|}{$0.85\times$ Body Scale}
& \multicolumn{3}{c}{$1.15\times$ Body Scale} \\
\cmidrule(lr){2-4} \cmidrule(lr){5-7}
& Succ. $\uparrow$ 
& $E_{\text{g-mpjpe}} \downarrow$
& $E_{\text{mpjpe}} \downarrow$
& Succ. $\uparrow$ 
& $E_{\text{g-mpjpe}} \downarrow$
& $E_{\text{mpjpe}} \downarrow$ \\
\midrule
PHC~\cite{luo2023perpetual}         & 0.756 & 116.795 & 79.844 & 0.703 & 143.521 & 107.418 \\
MaskedMimic~\cite{tessler2024maskedmimic} & 0.827 & 120.875 & 102.149 & 0.533 & 151.527 & 109.766 \\
Ours         & \textbf{0.879} & \textbf{92.827} & \textbf{64.614} 
             & \textbf{0.728} & \textbf{128.881} & \textbf{95.538} \\
\bottomrule
\end{tabular}
\end{table}


\section{Discussion}
\label{sec:discussion}
\begin{figure}[h] 
    \centering 
    \includegraphics[width=1\textwidth, trim=0cm 5cm 0cm 5cm, clip]{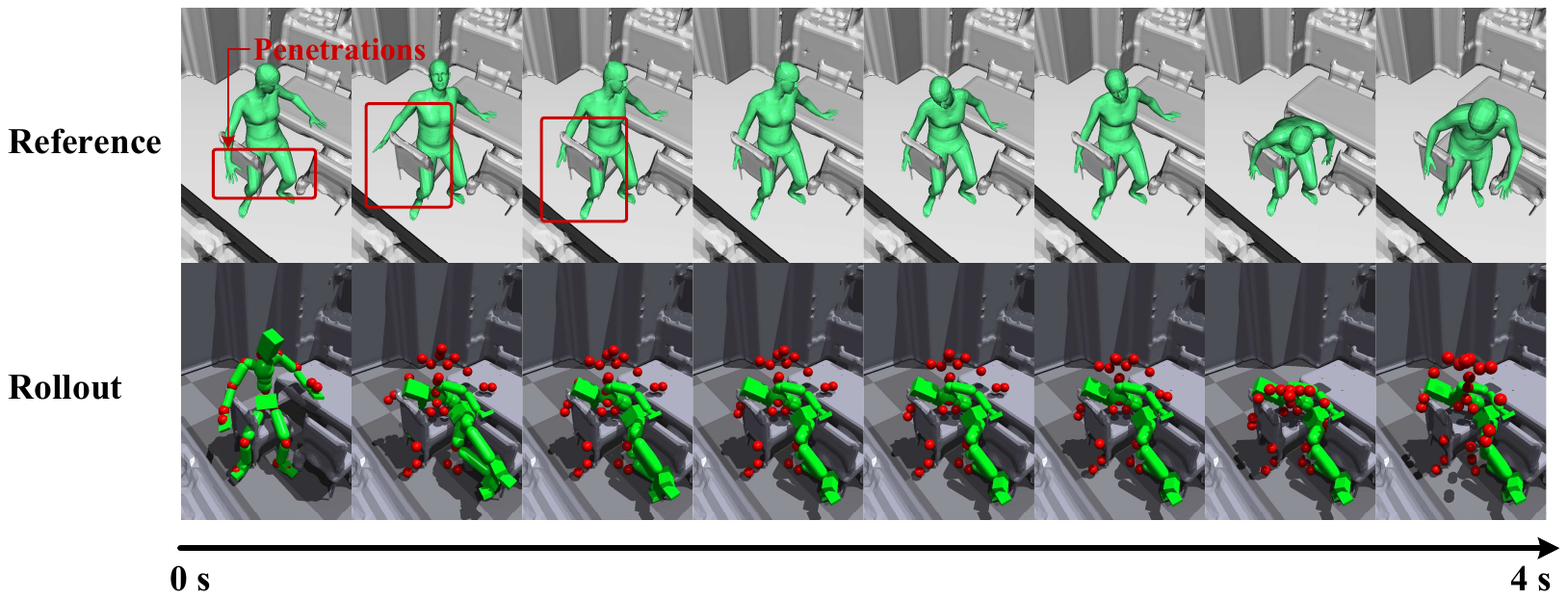}
    \caption{Examples of failure cases. The reference motion is severely penetrated with the 3D environments.}
    \label{fig:supp_figure_1} 
\end{figure}
\noindent\textbf{Limitations and Future Work.}
Although our method demonstrates strong performance on human–scene interaction imitation under complex 3D environments, it still has several limitations. First, our method cannot handle cases where the reference motion severely penetrates the scene meshes, as illustrated in \cref{fig:supp_figure_1}. Future work could explore improving the quality of the reference motions or incorporating penetration-aware filtering mechanisms. Second, similar to prior works on HSI~\cite{jiang2024scaling,hwang2025scenemi}, our framework focuses on body-level motion imitation and does not explicitly model fine-grained hand or finger interactions with objects. Incorporating articulated hand models and contact-aware control could further improve interaction realism.

\end{document}